\documentclass[runningheads]{llncs}
\usepackage{graphicx}
\usepackage{comment}
\usepackage{todonotes}
\usepackage{hyperref}
\usepackage{parskip}
\usepackage{xurl}

\begin{document}
%

\title{Scholarly Question Answering using Large Language Models in the NFDI4DataScience Gateway}

\titlerunning{Scholarly Question Answering}
%
\author{Hamed Babaei Giglou$^*$\inst{1}\orcidID{0000-0003-3758-1454} \and
Tilahun Abedissa Taffa\thanks{Equal contribution}\inst{2,3}\orcidID{0000-0002-2476-8335} \and
Rana Abdullah$^*$\inst{3}\orcidID{0009-0000-2652-5129} \and Aida Usmanova$^*$\inst{2}\orcidID{0009-0000-0124-8727} \and Ricardo Usbeck\inst{2}\orcidID{0000-0002-0191-7211} \and Jennifer D’Souza\inst{1}\orcidID{0000-0002-6616-9509} \and Sören Auer\inst{1}\orcidID{0000-0002-0698-2864}}

\authorrunning{Babaei Giglou et al.}

\institute{TIB Leibniz Information Centre for Science and Technology, Hannover, Germany \\\email{\{hamed.babaei,jennifer.dsouza,auer\}@tib.eu} \and \href{https://www.leuphana.de/en/institutes/iis/artifical-intelligence-and-explainability.html}{Artificial Intelligence and Explainability}, Leuphana Universität Lüneburg, Lüneburg, Germany\\ 
\email{aida.usmanova@stud.leuphana.de, ricardo.usbeck@leuphana.de}\and
Semantic Systems, Universität Hamburg, Hamburg, Germany
\\\email{\{tilahun.taffa,rana.abdullah\}@uni-hamburg.de}}
  
\maketitle              
\begin{abstract}

This paper introduces a scholarly Question Answering (QA) system on top of the NFDI4DataScience Gateway, employing a Retrieval Augmented Generation-based (RAG) approach. The NFDI4DS Gateway, as a foundational framework, offers a unified and intuitive interface for querying various scientific databases using federated search. The RAG-based scholarly QA, powered by a Large Language Model (LLM), facilitates dynamic interaction with search results, enhancing filtering capabilities and fostering a conversational engagement with the Gateway search. The effectiveness of both the Gateway and the scholarly QA system is demonstrated through experimental analysis.

\end{abstract}

\section{Introduction}

With recent advances in artificial intelligence (AI), decision-making has gradually shifted from rule-based systems to machine learning and deep learning-based developments~\cite{DBLP:conf/aaai/LahavSKJPSCYHWH22}. This paradigm shift has changed how we approach information retrieval and Question Answering (QA) systems, including Scholarly QA. Scholarly QA systems answer natural language questions over bibliographic data sources~\cite{auer2023sciqa,DBLP:conf/semweb/TaffaU23}. Notably, scholarly resources appear in different bibliographic repositories. To narrow down the answer search space - federated search comes into play. A federated search platform enables one to navigate the vast landscape of scholarly resources available across multiple databases and repositories~\cite{DBLP:journals/ftml/KairouzMABBBBCC21}.
Furthermore, federated search aggregates information from multiple sources to provide a comprehensive and holistic view of relevant resources. The efficacy of faceted search in scholarly-based filtering has been well-demonstrated~\cite{facetedsearch2020}, paving the way for robust systems employing federated search methods.

Adhering to FAIR principles~\cite{wilkinson2016fair} in managing research data, initiatives like the NFDI4DataScience\footnote{\url{https://www.nfdi4datascience.de/}} (NFDI4DS) consortium have emerged as a collaborative endeavor designed to support researchers throughout the entire research data life cycle, ensuring their practices align with the FAIR principles. The NFDI4DS Gateway, as a part of the NFDI4DS vision~\cite{DBLP:conf/gi/SchimmlerWBDKMK23}, includes a federated search. The Gateway - a unified and intuitive search interface that enables users to query various scientific databases such as DBLP, Zenodo, and OpenAlex. The overall aim of the NFDI4DS Gateway is to design an entry point that categorizes and summarises multiple search results (such as researchers, publications, machine learning models, and benchmark results) such that practitioners and researchers gain a swift overview of existing contributions~\cite{DBLP:conf/cordi/UsbeckAVASWCS23}.

Retrieval Augmented Generation (RAG) harnesses the power of advanced natural language processing (NLP) techniques to improve the quality and relevance of responses to user queries. Integrating Large Language Model (LLM)-based components is at the core of RAG's functionality. LLMs are the backbone of RAG's response generation process, leveraging extensive training on large text to understand and generate human-like responses. RAG-based scholarly QA systems can seamlessly integrate with federated search to improve the process of filtering and selecting scholarly resources in the context of scholarly research~\cite{DBLP:journals/corr/abs-2312-07559}. 
Therefore, on top of the NFDI Gateway, we built a RAG-based scholarly QA system. The RAG retrieval component of the system scans the retrieved resources to identify the top-N most relevant documents based on the user's question. After placing the resources, the scholarly QA uses an LLM (Large Language Model) to extract correct answers to the user's questions directly from the selected documents. By seamlessly integrating the RAG-based scholarly QA with the Gateway, users can efficiently filter through vast amounts of scholarly content, enabling more targeted and productive research efforts.

This approach aims to enhance the user experience, fostering more intuitive and tailored engagement with available information, ultimately contributing to more effective and nuanced research outcomes. Furthermore, a detailed analysis of our experiments is framed as two main research questions (RQs). 

\begin{itemize}
    \item \textbf{RQ1:} To what extent does the federated search implemented in the Gateway achieve optimal performance?
    \item \textbf{RQ2:} How does integrating the Scholarly QA on top of the Gateway improve the retrieval of relevant search results?
\end{itemize}

Our main contributions are twofold:

\begin{itemize}
    \item the NFDI4DS Gateway analysis and completeness of federated search evaluation through information retrieval metrics.
    \item A scholarly QA system based on RAG on top of the Gateway.

\end{itemize}

The source code can be found at \url{https://github.com/semantic-systems/nfdi-search-engine-chatbot}.

\section{Related Works}

Data management is a multi-step process that involves obtaining, cleaning, and storing data to allow accurate analysis and produce meaningful results. As an example of this, the Open Research Knowledge Graph~\cite{orkg23,orkg20} is an infrastructure for the production, curation, publication, and use of FAIR scientific information with the ultimate goal of providing swift knowledge management within the scientific domain by the digitalization of scholarly articles in the form of the knowledge graph. On the other hand, the federated search~\cite{DBLP:series/lncs/YangTZCY20}, as they involve the efficient retrieval of information from multiple data sources, play an essential role in data management as it helps in optimizing the use of data and deriving valuable insights from the data. As shown in~\cite{DBLP:conf/wims/AuerKPKSV18,DBLP:conf/aaai/LahavSKJPSCYHWH22} work, the researchers face a flood of papers that hinders the discovery of necessary knowledge, as a result of this, \cite{DBLP:conf/aaai/LahavSKJPSCYHWH22} trained models to identify challenges and directions across the corpus by a dedicated search engine. 

Federated search~\cite{DBLP:journals/ftir/ShokouhiS11} serves as a crucial tool for managing data within scholarly articles, enabling the retrieval of information from diverse sources through a search application constructed on top of one or more data sources~\cite{DBLP:journals/ftml/KairouzMABBBBCC21}. A federated search facilitates information retrieval from multiple scholarly sources, demonstrating remarkable efficacy across various fields, particularly in scientific data management. Shokouhi et al.~\cite{DBLP:journals/ftir/ShokouhiS11} outlined the challenges inherent in federated search within scholarly domains, delineating three significant hurdles: retrieving relevant documents, identifying suitable collections necessitating knowledge representation and unifying results from multiple sources. 
Similarly, Kumar et al.~\cite{Kumar2007} dived into how federated search helps libraries and other institutions with a valuable tool to explore various fields and articles. Furthermore, Kirstein et al.~\cite{kirstein2020piveau} introduced \textit{Piveau} as a comprehensive open data management solution grounded in semantic web technologies. Leveraging a spectrum of standards prevalent in the semantic web, such as RDFs and DCAT, this standardization via the semantic web overcomes limitations in search capabilities, ensuring superior quality information retrieval.


The Scholarly QA work in~\cite{DBLP:journals/corr/abs-2312-07559} proposes a QA model that extracts question-related full-text scientific articles using an LLM-based retrieval agent and generates answers using RAG techniques. ~\cite{DBLP:conf/semweb/TaffaU23} has explored Knowledge Graph QA using an LLM in a few-shot setting for handling bibliographic questions. NLQxform~\cite{DBLP:journals/corr/abs-2311-07588} introduces a natural language interface for directly querying the DBLP by automatically translating questions into SPARQL queries. Unlike these Scholarly QAs, we introduce RAG-based Scholarly QA.

\section{Methodological Framework}
\begin{figure}[tb]
\includegraphics[width=\textwidth]{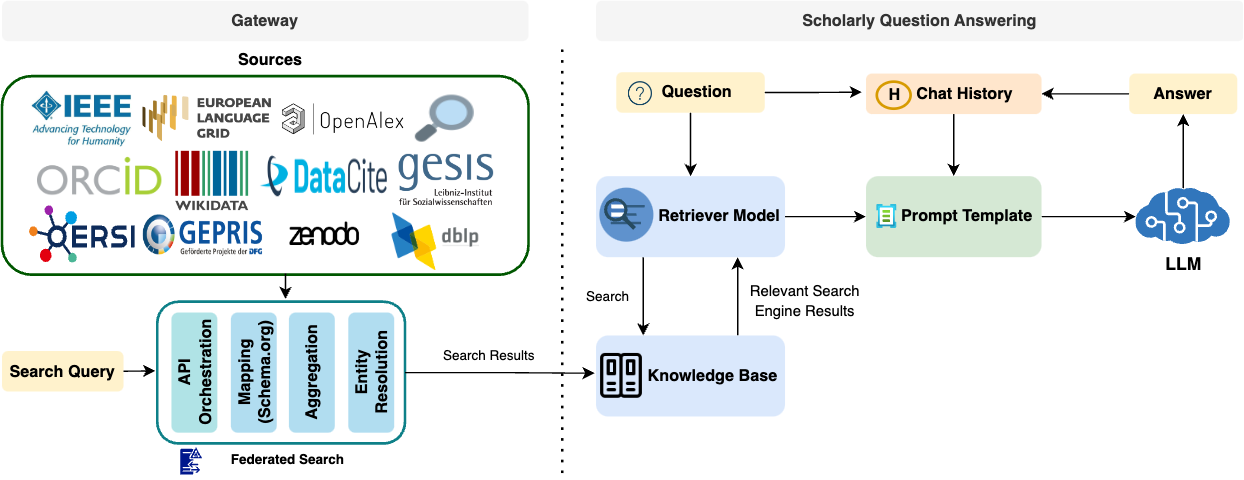}
\caption{A functional view of the NFDI4DS Gateway architecture with scholarly QA application.} \label{archnfdi4dsarch}
\end{figure}

The NFID4DS Gateway performs a federated search through various data store APIs. Subsequently, the search results will be indexed into the QA system to allow users to find acquired information via chat. The architectural representation of the Gateway with a scholarly QA system is illustrated in~\autoref{archnfdi4dsarch}. 

\subsection{The Gateway -- Federated Search}
The Gateway conducts federated searches across diverse data stores, generating results that humans can easily interpret. The following key components underpin its functionality:
\begin{enumerate}
    \item Keyword search across data stores (\textsc{API Orchestration}).
    \item Grouping results using a faceted taxonomy (\textsc{Mapping and Aggregation}).
    \item Deduplication of results (\textsc{Entity Resolution}).
\end{enumerate}
In the following, we will delve into each of these components in detail, explaining their functions and contributions to the overall functionality of the Gateway.

\noindent\textbf{1) API Orchestration.} It uses a one-search-box interface to obtain user keywords, and the search results are expressed in a one-result-list-only manner. It subsequently employs federated search using ad-hoc based searches through 11 open-source scholarly repositories, i.e., DBLP\footnote{\url{https://dblp.org/}}, OpenAlex\footnote{\url{https://openalex.org/}}, CORDIS\footnote{\url{https://cordis.europa.eu/}}, European Language Grid (ELG)\footnote{\url{https://live.european-language-grid.eu/}}, GEPRIS\footnote{\url{https://gepris.dfg.de/}}, GESIS\footnote{\url{https://www.gesis.org/en/home}}, ORCiD\footnote{\url{https://orcid.org/}}, RESODATE\footnote{\url{https://resodate.org/}}, WIKIDATA\footnote{\url{https://www.wikidata.org/}}, IEEE\footnote{\url{https://www.ieee.org/}}, and Zenodo\footnote{\url{https://zenodo.org/}}. Among these repositories, DBLP, OpenAlex, IEEE, GESIS, RESODATE, WIKIDATA, and Zenodo provide research resources like publications, datasets, software, etc. GEPRIS provides Deutsche Forschungsgemeinschaft (DFG) funded projects. Likewise, CORDIS is a primary source for projects financed by the European Union (EU) Commission. ELG is a platform that provides multi-lingual, cross-lingual, and mono-lingual language technologies in the EU. Unlike the others,  ORCiD delivers a unique, persistent, researcher-owned, and controlled digital identifier that distinguishes researchers uniquely.

\noindent\textbf{2) Mapping and Aggregation.} The Gateway interacts with various data source APIs, including SPARQL endpoints; the retrieved results often have different structures. For example, while one source refers to an author of a publication as `author', another refers to them as `creator', and terminology differences extend to scholarly resources such as datasets, which are referred to as `corpus' in one source and `dataset' in another, such as Zenodo. To resolve the naming variations, we have developed a systematic approach based on customized faceted taxonomy from schema.org\footnote{\url{https://schema.org/}} that harmonizes and aggregates heterogeneous results from API orchestration. This faceted taxonomy acts as a unifying framework that allows us to map the different terminology and structures found in other data sources, thereby coherently facilitating the aggregation and presentation of search results.

The faceted taxonomy based on schema.org is defined to represent different entities found in data sources. These schema.org classes encompass information including organizations, individuals, authors, creative works (articles, datasets, projects, software applications, learning resources, and media objects), and their respective attributes. In particular, the Author and Person classes encapsulate attributes related to individuals who contribute to creative works, while the Organization class encapsulates attributes specific to organizational entities. In addition, the \texttt{CreativeWork} super class serves as a foundation for various entities, providing common attributes such as \texttt{abstract}, \texttt{author}, and \texttt{datePublished} inherited by its subclasses. Each class within schema.org contributes to a structured representation of data entities, facilitating organization, interoperability, and standardized data handling within the Gateway.

\noindent\textbf{3) Entity Resolution.} Following the initial mapping of the publications, researchers, and other resources using the schema.org taxonomy, it becomes necessary to identify and merge duplicate objects within the results. To accomplish this task, we leverage the  DEDUPE model~\cite{Greggdedupe2022}, which employs machine learning techniques, specifically fuzzy matching, deduplication, and entity resolution, to handle structured data effectively. Later, the DEDUPE model can be fine-tuned on a custom dataset comprising positive and negative samples, thus enabling the model to differentiate between genuine duplicates and distinct entities. 

For publication deduplication, the DEDUPE model is trained on a set of attributes, i.e., Digital Object Identifier (DOI), title, author list, abstract, and publication date for publication identification by clustering objects based on similarity scores calculated across attributes. Subsequently, within each cluster, objects that exceed the predefined similarity threshold are merged to form a unified entity, thus resulting in a set of deduplicated records. Later, the resulting records are sorted based on relevancy score using BM25Plus. BM25Plus is a variant of BM25 (Best Match)~\cite{DBLP:journals/ftir/RobertsonZ09} ranking algorithm, introducing additional term weighting factors to enhance the ranking.

\subsection{Scholarly Question Answering}
As shown in~\autoref{archnfdi4dsarch}, our RAG-based~\cite{DBLP:conf/nips/LewisPPPKGKLYR020} scholarly QA has two components: (i) a retriever that returns top-K relevant passage to the user's question and (ii) a generator LLM that generates a human-like response based on a given context from the retriever to a user question. 

\noindent\textbf{Retriever.} The retrieval model uses a user question as a query to explore relevant information from a knowledge base. The knowledge base comprises a set of documents retrieved per search query through the Gateway. The retriever model operates in three sequential steps:
\begin{enumerate}
    \item \textsc{Step 1}: The preprocessing knowledge base of search results to obtain a set of documents combined textual data by combining the key-value dictionary per obtained search result.
    \item \textsc{Step 2}: The retriever model extracts embeddings for the documents and indexes them within the knowledge base. 
    \item \textsc{Step 3}: Given a specific question, the retriever model extracts embeddings and computes cosine similarity with the knowledge base, thereby retrieving the top-K appropriate relevant documents to answer the question
\end{enumerate}
We opted for an ensemble retriever model. This ensemble accompanies techniques such as TF-IDF~\cite{DBLP:reference/ml/X10vu}, SVM, and KNN retrievers with the Sentence-BERT~\cite{DBLP:conf/emnlp/ReimersG19} model serving as the foundational framework. Per the user question, the ensemble retriever queries retriever models to obtain their results; next, it ranks them using each retriever's weights to obtain the final documents most similar to the query. In our retriever collection, the SVM is being trained with the query as a positive class and the rest of the knowledge base documents as negative using sentence-BERT embeddings; next, based on the positive class probability, the documents are ranked and obtain top-k items. By integrating diverse retrieval methodologies, our ensemble model aims to capitalize on the strengths of each component, thereby enhancing overall retrieval performance. Upon experimentation, we manually determined the optimal configuration for our ensemble model. Based on our observations, we assigned weights of 0.3 to TF-IDF, 0.3 to KNN, and 0.4 to SVM retrievers by try-and-error analysis. 

\noindent\textbf{Generator.} As shown in~\autoref{archnfdi4dsarch}, the generator model uses LLM and retriever documents and a prompt template to query LLM to generate a human-like answer to the user questions based on obtained relevant documents from the retriever model. As observed, LLMs showed a great capability for generating human-like responses. However, they might hallucinate and forget the discussion due to the overwhelming information. We provide explicit instructions beside questions and relevant documents, using a predefined prompt template to avoid this. The prompt template enables the scholarly QA to query LLM effectively and answer the user question accurately. The prompt template is described as follows:

\begin{quote}
    \small
    \textcolor{blue}{
    Provide your answers only on the knowledge provided here. Do not use any outside knowledge. \\
    If you don't know the answer, say that you don't know. Don't try to make up an answer. \\
    \bigskip
    Given the following context, answer the below question: \\    
    \bigskip
    \textcolor{red}{\{context\}} \\    
    Question: \textcolor{red}{\{question\}} \\
    Helpful Answer:
    }
\end{quote}
In the prompt template, \textit{\textcolor{red}{\{context\}}} is the placeholder for retriever model results, and \textit{\textcolor{red}{\{question\}}} is the user question. To account for follow-up questions, we have used conversation buffer memory that keeps track of chat history, consisting of previous questions and answers within five previous conversations. The follow-up questions can reference past chat history, e.g., ``What is the open research knowledge graph?" followed by ``How to use it?" Such queries challenge direct retriever similarity-based searches, including ensemble retriever models. We provided the chat history for LLM in the prompt template by adding the history questions and answers to the end of retrieval model outputs at \textit{\textcolor{red}{\{context\}}} placeholder. As an LLM, we use GPT-3.5~\cite{chatgpt} with the LangChain framework~\cite{ChaseLangChain2022} for implementation.

\section{Evaluation}

\subsection{Evaluation Dataset}
This section outlines the procedures for constructing the dataset for both the Gateway and scholarly QA evaluations. 

\subsubsection{Constructing Queries for Assessing the Gateway Performance.} 

The comparison feature of ORKG empowers researchers to construct comprehensive comparisons~\cite{stocker2023fair} among scholarly articles spanning diverse domains. A pivotal aspect of this feature is the inclusion of human-generated comparisons. In the evaluation of federated search, we focused on the comparison titles at ORKG, crafted by the researchers themselves. Consider a scenario where a user aims to formulate a comparison for ``ontology learning from text" and utilizes the Gateway to gather relevant papers and sources for their study. When a user queries the title on the Gateway, a user can easily use the documents obtained to construct an ORKG comparison for ``ontology learning from the text" as shown in \href{https://orkg.org/comparison/R186047/}{https://orkg.org/comparison/R186047}. So, comparison titles can be used as a query to study the Gateway's performance in finding relevant documents for researchers. 

Through this process, we obtained 1,235 unique comparisons from ORKG as of \textit{February 2nd, 2024}, spanning 161 research fields. Among the obtained research fields, we selected 27 research fields related to AI and data science. Consequently, we identified 316 comparison topics within 27 research fields that fall into the AI and data science category for human annotations to curate titles as a query. Ultimately, we curated a collection of 275 comparison titles for performance analysis of the Gateway and executed queries on the Gateway as of \textit{February 16th, 2024}. The remaining 41 comparison titles we found them inappropriate for querying the Gateway.

\subsubsection{Generating Scholarly QA Datasets.} 

We designed a systematic approach to generate well-suited questions tailored to search results. The questions are designed to simulate what questions users ask while using the Gateway. We constructed the AI-QA dataset using GPT-4~\cite{DBLP:journals/corr/abs-2303-08774} and the Comparison-QA dataset using ORKG comparisons. For the AI-QA dataset, we employed k-means~\cite{DBLP:reference/ml/JinH10a} clustering methodology on retrieved documents per query, enabling us to efficiently organize the data for generating questions. For search result sets containing more than 50 entries, we applied a clustering number of 10, and for result sets with fewer than 50 entries, a clustering number of 5 was considered appropriate. Search results with less than 5 entries were not included in question generation. Subsequently, we employed GPT-4-Turbo~\cite{DBLP:journals/corr/abs-2303-08774} to generate two appropriate questions per cluster using a predefined prompt template. The prompt template is defined as follows:

\begin{quote}
\textcolor{blue}{The task is to generate questions based on the provided information.\\
Given a list of texts, generate only two questions, no more than two.\\
Make questions variant.\\
The questions should imitate what a user might look for in the given documents.\\
\\
Return questions as a Python list.\\
\\
Documents:\\
\textcolor{red}{\{documents\}}}
\end{quote}
This approach proves advantageous in generating questions for scholarly QA evaluation as it relies on documents already recognized for question generation. However, in the evaluation phase, the retriever model gathers search results similar to those of the questions, which the LLM later uses to generate answers. Following the question generation step, we acquired a total of \textbf{3,298} questions across 1,651 clusters for scholarly QA evaluations, where we consider each clustering per question as a ground truth.
\newline
\newline
Since the ORKG comparison is aimed to allow researchers to compare contributions of different articles based on predefined properties such as ``research problem" or ``model". For Comparision-QA, we used comparison properties as questions using the following standard template:
\begin{quote}
    \textcolor{blue}{In the paper ``\textcolor{red}{\{paper\}}", what is the \textcolor{red}{\{property\}}?}
\end{quote}
We considered 275 comparison titles to query the Gateway to obtain federated search results; for the 275 ORKG comparisons comprising 2,395 papers, only 184 were retrieved by the Gateway. So, we used 184 papers and their properties to construct questions, and values for the property per paper in the comparison were considered as answers. In the end, a total of 1,354 questions were constructed.

The overview of the datasets is presented in \autoref{tab:statistics}.
\begin{table}[]
    \centering
    \caption{Statistics for the number of search queries (Query), number of comparison papers (Comparison Papers), number of papers from ORKG comparison that are being covered in search results (ORKG Coverage), and comparison specific questions (Comparison-QA).}
    \label{tab:statistics}
    \begin{tabular}{|r|r|r|r|r|}
          \hline
         \textbf{Query}&  \textbf{AI-QA}  & \textbf{Comparison Papers} & \textbf{ORKG Coverage} & \textbf{Comparison-QA}\\
         \hline
         275 &  3,298 & 2,303 & 184&1,354\\
         \hline
    \end{tabular}
\end{table}

\subsection{Evaluation Metrics}
\subsubsection{Gateway Evaluation Metrics.} In evaluating the performance of the Gateway, we employed multiple approaches, primarily focusing on response time, number of retrieved documents, and relevancy scores. The response time analysis serves as a critical metric in assessing the efficiency and responsiveness of the Gateway. Another key aspect of our evaluation involved analyzing the number of documents retrieved by the Gateway in response to user queries. This metric provides valuable information about the comprehensiveness and effectiveness of the search results generated by the system. To further refine our evaluation, we calculated relevancy scores per retrieved document similarity to the search query based on varying thresholds and representations such as sentence-BERT, TF-IDF, and BM25~\cite{DBLP:reference/db/Amati09}. With sentence-BERT sentence embeddings, TF-IDF, and BM25 scores, we calculated cosine similarity between documents and queries for all metrics to get relevancy scores.

\subsubsection{Scholarly QA Evaluation Metrics.} In AI-QA, we utilized question clusters as answers, while in comparison-QA, property values were employed as answers. Subsequently, we assessed performance using n-gram overlap specific metrics like ROUGE~\cite{lin-2004-rouge} (Recall-Oriented Understudy for Gisting Evaluation) and BLEU~\cite{papineni-etal-2002-bleu} (Bilingual Evaluation Understudy), focusing specifically on ROUGE-1, ROUGE-L, and BLEU-1 as our evaluation criteria. Because LLMs generate responses based on their comprehension, they might deviate from the ground truth text, making evaluation with metrics like ROUGE and BLEU difficult. Consequently, incorporating similarity scores into the assessment process can offer further insights into their proficiency in capturing subtle language nuances. We used the BERTScore -- a sentence-BERT average cosine similarity metric as an evaluation. Furthermore, as the Comparison-QA dataset poses challenges with answers often appearing within the paper context rather than solely in abstracts and titles, we opted for the Exact Match score as another evaluation metric only for this dataset.

\subsection{Results}




\subsubsection{Gateway and Scholarly QA Results.} 
\begin{figure}[ht]
    \centering
    \includegraphics[width=\textwidth]{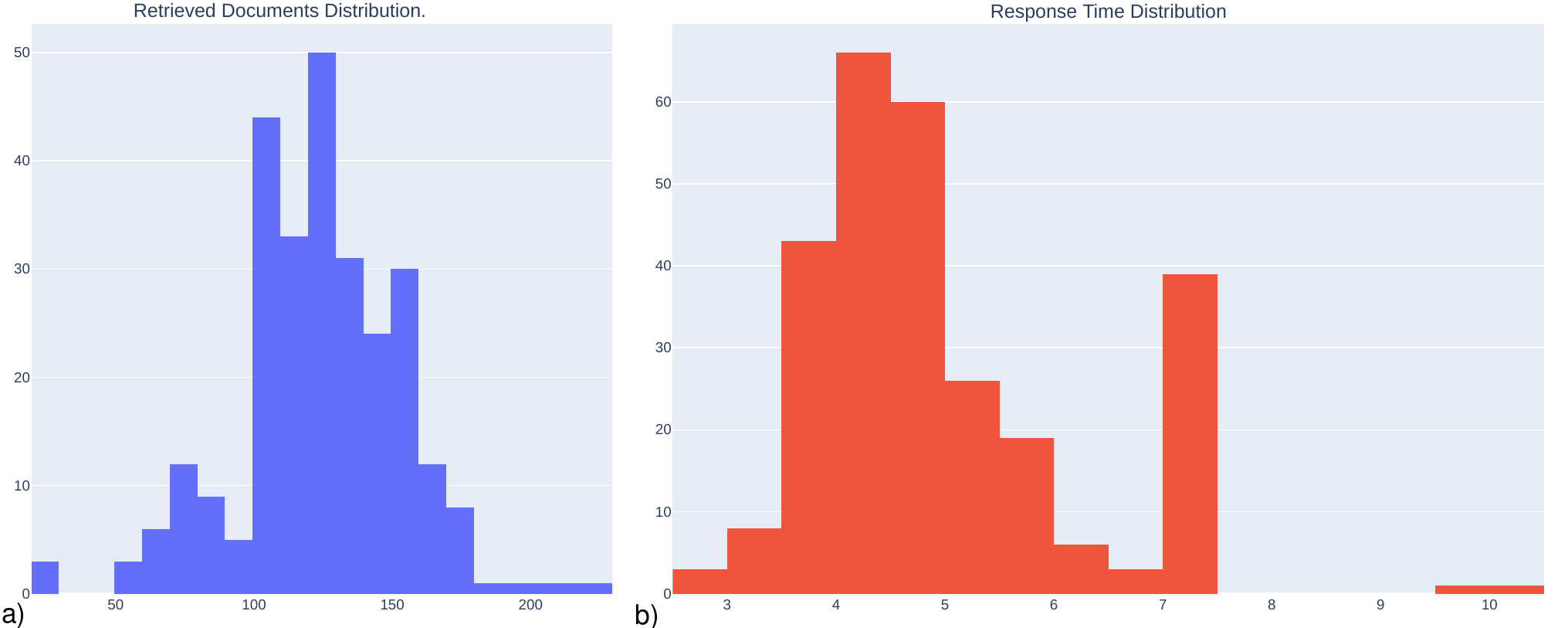}
    \caption{Gateway retrieved documents distribution is presented in the left figure. The x-axis represents the number of retrieved documents, and the y-axis the number of queries. The right figure represents the response time distribution, with the x-axis as a response time in seconds and the y-axis as the number of queries.}
    \label{fig:response_time_results}
\end{figure}

\begin{figure}[ht]
    \centering
    \includegraphics[width=\textwidth]{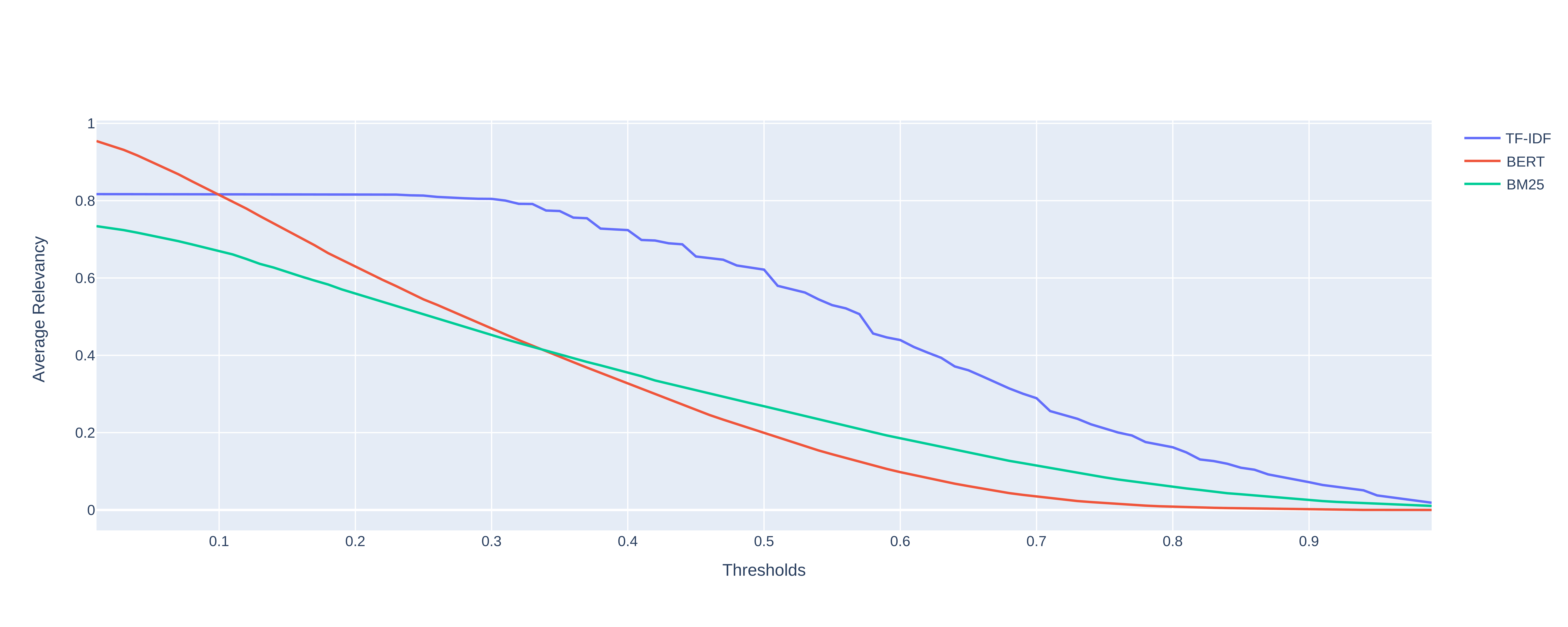}
    \caption{Gateway retrieved documents relevancy w.r.t search query analysis using TF-IDF, BM25, and sentence-BERT embeddings for similarity measurement and different thresholds in the range of [0.0, 0.99]}
    \label{fig:relevancy}
\end{figure}

The performance of the Gateway has been assessed by considering factors such as its response time, the number of documents retrieved, and the relevance of those documents. The Gateway performances are reported in~\autoref{fig:response_time_results} and~\autoref{fig:relevancy}.
The results for scholarly QA evaluation, employing various metrics, are reported in~\autoref{tab:chatbot_result}. We identified 432 questions without answers for AI-QA, while we obtained 26 questions without answers for Comparison-QA. This happened due to the input limitation of GPT-3.5. Hence, we excluded these questions from evaluations.

\begin{table}[ht]
    \centering
    \caption{Evaluation results of the scholarly QA using AI-QA and Comparison-QA datasets, showcasing ROUGE, BLEU, BERTScore, and Exact Match scores for the RAG-based scholarly QA development.}
    \label{tab:chatbot_result}
    \begin{tabular}{|l|r|r|r|r|r|r|}
        \hline
         \textbf{Dataset} & \textbf{ROUGE-1} & \textbf{ROUGE-L} & \textbf{BLEU-1}  & \textbf{BERTScore} & \textbf{Exact Match}  \\
         \hline
         \textit{AI-QA} & 4.21  & 2.92  &  38.94 & 36.81 & - \\
         \textit{Comparison-QA} & 6.82 & 6.10 & 3.10 & 26.96 & 13.93\\
         \hline
    \end{tabular}
\end{table}

\noindent\textbf{RQ1: [Gateway] To what extent does the federated search implemented in NFDI4DS achieve optimal performance?} We address this question by analyzing the findings presented in~\autoref{fig:response_time_results} and~\autoref{fig:relevancy}. Ultimately, for a search platform, it is essential to retrieve relevant results while maintaining a fast response time across various queries. The analysis of response time and retrieved documents status in~\autoref{fig:response_time_results} for 275 search queries showed that the federated search is capable of obtaining \textbf{123} documents on average within an average response time of \textbf{4.93} seconds. Notably, slow performance is observed in the search query of the ``Kinect human activity recognition dataset" with approximately 10 seconds response time and search results of 169 documents.
Similarly, for the ``Motion Capture system" search query, we obtained 227 documents within 4.3 seconds. This shows that depending on different search keywords and how complex the query is, it may result in sacrificing response time. In general, according to~\autoref{fig:response_time_results}, the distribution analysis indicated that the number of retrieved documents follows a \textit{normal} distribution, while the distribution of response time is \textit{positively skewed}. This highlights the significant performance of the Gateway in terms of response time and document retrieval.
\newline
We calculated cosine similarities with three metrics to analyze the retrieved documents' relevancy. We set relevancy thresholds to see how many queries with their corresponding documents are considered very relevant to each other.
The relationship between the relevancy threshold and the number of retrieved documents is depicted in~\autoref{fig:relevancy}, indicating a decrease as the threshold increases. The TF-IDF metric generates the highest similarity scores between documents and queries, albeit focusing primarily on token frequency rather than semantic understanding. BM25, an improvement upon TF-IDF, proves particularly effective for information retrieval tasks, displaying a different score distribution with numerous low similarity scores. Despite this, BM25 still identifies certain documents as highly relevant (with similarity above 0.3) for specific queries. Conversely, sentence-BERT initially achieves the highest average recall but drops to zero at a threshold of 0.8. Comparatively, BM25 and sentence-BERT yield similar results, implying that capturing nuanced semantics may not be crucial for retrieving relevant articles; instead, identifying standard terms and phrases appears more pivotal. Evaluating the optimal threshold of 0.3, TF-IDF emerges as the optimal ranking model. The overall relevancy analysis across different thresholds indicates that the Gateway effectively retrieves search results based on keyword search but struggles with semantic retrieval. However, setting the threshold to 0.3 demonstrates approximately 50\% semantic similarity among documents, highlighting the Gateway's proficiency in identifying relevant documents from keyword and semantic perspectives.

\noindent\textbf{RQ2: [Scholarly QA] How does integrating the Scholarly QA on top of the Gateway improve the retrieval of relevant search results?} We address this question by analyzing the results presented in~\autoref{tab:chatbot_result} for both automated constructed Comparison-QA and AI-QA datasets. According to the ROUGE-1 metric, unigrams overlap between the developed QA-generated responses and existing answers. This overlap is more significant for Comparison-QA (6.82\%) than for AI-QA (4.21\%). Similarly, when considering ROUGE-L, which measures the Longest Common Subsequence, the overlap for Comparison-QA (6.10\%) surpasses that of AI-QA (2.92\%). However, despite the QA's promising BLEU-1 score of 38.94\% on the AI-QA dataset, its performance on the Comparison-QA dataset is lacking. This suggests that the developed QA responses align more closely with the clustered documents, which are the ground truth in our AI-QA dataset. 

It is essential to note that both the ROUGE and BLEU metrics have limitations when applied to LLM-based generations. This is because LLM-generated responses may exhibit variations that mimic human-like responses, making it challenging for these metrics to evaluate their quality accurately. Still, they show how much of the generated text is similar to ground truth. Nevertheless, we reported a BERTScore of 36.81\% for the AI-QA dataset and 26.96\% for the Comparison-QA dataset. These obtained BERTScore results suggest that the quality of the scholarly QA's responses, particularly in terms of semantic similarity to ground truth references, varies significantly between the two datasets. As mentioned earlier, the variation between the two datasets was expected since the Comparison-QA mostly extracted humans from the whole body of the paper rather than only the title and abstract. 

We computed the exact match for Comparison-QA, revealing a 13.93\% match between the ground truth and the QA-generated text. This highlights the scholarly QA's proficiency in recognizing relevant information, mainly when it appears in the search results. To the best of our knowledge, there is no other baseline system or scholarly QA system available to which we can compare.

\section{Limitations and Future Directions}
This section discusses the limitations encountered in the implementation of the scholarly QA model and outlines potential future directions for addressing these shortcomings. 

\noindent\textbf{Inadequate Availability of Comparison-QA Dataset Answers.} The scholarly QA's performance is hindered by the frequent unavailability of answers to the Comparison-QA answers in search results, resulting in suboptimal performance. Addressing this limitation requires an extensive collection of queries from ORKG comparisons. Another limitation arises from the lack of diversity in the questions, as the current methodology employs a single template for forming questions on this dataset.

\noindent\textbf{Suboptimal AI-QA Dataset Generation.} The AI-QA dataset, generated from clustered search results, sometimes yields many documents per cluster. Thus, an optimal clustering method is necessary to manage the data effectively. Additionally, soliciting human feedback on the generated questions is crucial for refining and enhancing the dataset's quality. In future works, it is helpful to have a small human-generated dataset to justify the evaluation's validity further.

\noindent\textbf{Exploring Diverse LLMs.} Future research should focus on exploring a more comprehensive range of LLMs within scholarly QA to study their diversity and identify more optimal models for scholarly documents. This endeavor necessitates dataset curation tailored explicitly to the Gateway results.

\section{Conclusion}
In this work, we present an interactive scholarly QA system based on the RAG approach on top of the NFDI4DataScience Gateway search results, facilitating user interaction with a wealth of data. Subsequently, we automatically evaluated both the Gateway and scholarly QA using an automatically constructed dataset. The analysis indicates that as early prototypes, both the Gateway and QA show satisfactory performance. However, there is a need for future work to stabilize both systems and harness data science expertise.

\noindent\subsubsection{Acknowledgments}
This work was jointly supported by the German Research Foundation (DFG) project NFDI4DS under Grant No.: 460234259 and the German BMBF project SCINEXT under grant No. 01lS22070.

%
%
%
\bibliographystyle{splncs04}
\bibliography{mybibliography}

\end{document}